\documentclass{article}
\usepackage{ijcai20}

\usepackage{times}
\usepackage{soul}
\usepackage{url}
\usepackage[hidelinks]{hyperref}
\usepackage[utf8]{inputenc}
\usepackage[small]{caption}
\usepackage{graphicx}
\usepackage{amsmath}
\usepackage{amsthm}
\usepackage{booktabs}
\usepackage{algorithm}
\usepackage{algorithmic}
\newtheorem{theorem}{Theorem}

\usepackage{dblfloatfix}
\newcommand{\ourmodel}[0]{HypE}
\newcommand{\oursimple}[0]{HSimplE}
\newcommand{\acr}[1]{{{\textsc{#1}}}}
\usepackage{subcaption}
\usepackage{verbatim}
\newcommand{\relset}[0]{{\set{R}}}
\newcommand{\entset}[0]{{\set{E}}}
\newcommand{\maxar}[0]{\alpha}
\newcommand{\set}[1]{\ensuremath{\mathcal{#1}}}
\newcommand{\realset}[0]{\ensuremath{\mathbb{R}}}
\newcommand{\dotsum}[1]{\ensuremath{\odot(#1)}}
\newcommand{\defeq}{\ensuremath{=}}
\newcommand{\eg}[0]{\emph{e.g.},~}
\newcommand{\tuplee}[1]{\ensuremath{\odot(#1)}}
\newcommand{\tuplew}[4]{\ensuremath{#1(#2, #3, #4)}}
\newcommand{\rank}[0]{\func{rank}}
\newcommand{\func}[1]{\ensuremath{\mathrm{#1}}}
\usepackage{booktabs}       
\usepackage{caption}
\newcommand{\vc}[1]{\ensuremath{\mathbf{#1}}}
\newcommand{\tupleexample}{\ensuremath{\tuplew{r}{e_i}{\dots}{e_j}}}
\usepackage{amsmath}
\usepackage{amsfonts}
\usepackage{mathtools}
\DeclarePairedDelimiter\floor{\lfloor}{\rfloor}
\newcommand{\entityv}[0]{\vc{e}}
\usepackage{amsmath,amsfonts,amsthm} 
\newcommand{\ie}[0]{\emph{i.e.},~}
\usepackage{bm}
\newenvironment{hproof}{%
  \proof}{\endproof}
\usepackage{textcomp}

\title{Knowledge Hypergraphs: Prediction Beyond Binary Relations\thanks{This paper with the supplementary material can be found at \href{https://arxiv.org/abs/1906.00137}{https://arxiv.org/abs/1906.00137}.}}

\author{
Bahare Fatemi$^{1,2}$\footnote{Contact Author}\and
Perouz Taslakian$^2$\and
David Vazquez$^2$\And
David Poole$^1$\\
\affiliations
$^1$University of British Columbia\\
$^2$Element AI\\
\emails
\{bfatemi, poole\}@cs.ubc.ca,
\{perouz,dvazquez\}@elementai.com,
}

\begin{document}

\maketitle

\begin{abstract}
Knowledge graphs store facts using relations between two entities. In this work, we address the question of link prediction in knowledge hypergraphs where relations are defined on \emph{any number} of entities. While techniques exist (such as reification) that convert non-binary relations into binary ones, we show that current embedding-based methods for knowledge graph completion do not work well out of the box for knowledge graphs obtained through these techniques. To overcome this, we introduce \emph{\oursimple{}} and \emph{\ourmodel{}}, two embedding-based methods that work directly with knowledge hypergraphs. In both models, the prediction is a function of the relation embedding, the entity embeddings and their corresponding positions in the relation. We also develop public datasets, benchmarks and baselines for hypergraph prediction and show experimentally that the proposed models are more effective than the baselines. 
\end{abstract}

\section{Introduction}
\label{sec:introduction}

\emph{Knowledge hypergraphs} are graph structured knowledge bases that store facts about the world in the form of relations among any number of entities. They can be seen as one generalization of \emph{knowledge graphs} in which relations are defined on at most two entities. 
Since accessing and storing all the facts in the world is difficult, knowledge bases are incomplete; the goal of \emph{link prediction} in knowledge (hyper)graphs (or \emph{knowledge (hyper)graph completion}) is to predict unknown links or relationships between entities based on existing ones. 
In this work, we are interested in the problem of link
prediction in knowledge hypergraphs. Our motivation for studying link prediction in these more sophisticated knowledge structures is based on the fact that most knowledge in the world has inherently complex compositions.

Link prediction in knowledge graphs is a problem that is studied extensively,
and has applications in several tasks such as
automatic question answering~\cite{watson}. 
In these studies, knowledge graphs are defined as directed graphs having nodes as entities and labeled edges as relations; edges are directed from the \emph{head} entity to the \emph{tail} entity. 
The common data structure for representing knowledge graphs is a set of triples $relation(head, tail)$ that represents information as a collection of binary relations. 
There exist a large number of knowledge graphs that are publicly available, such as 
\acr{Freebase}~\cite{bollacker2008freebase}.
\citeauthor{m-TransH} (\citeyear{m-TransH}) observe that in the original \acr{Freebase} more than $1/3$rd of the 
entities participate in non-binary relations (i.e., defined on more than two entities). We observe, in addition, that $61$\% of the relations in the original Freebase are non-binary. 

Embedding-based models~\cite{nguyen2017overview,survey-graph} have proved to be effective for knowledge graph completion. These approaches learn embeddings for entities and relations.
To find out if a triple $relation(head, tail)$ is true, such models define a function that uses embeddings of $relation$, $head$, and $tail$ and output the probability of the triple.
While successful, such embedding-based methods make the strong assumption that all relations are binary.

Knowledge hypergraph completion is a relatively under-explored area. We motivate our work by outlining that converting non-binary relations into binary ones using methods such as reification or star-to-clique~\cite{m-TransH}, and then applying known link prediction methods does not yield satisfactory results.
\emph{Reification} is one common approach of converting higher-arity relations into binary ones.
In order to reify a tuple having a relation defined on~$k$ entities $e_1, \dots, e_k$, 
we form~$k$ new binary relations, one for each position in this relation, and a new entity $e$ for this tuple
and connect $e$ to each of the $k$ entities that are part of the given tuple using the $k$ binary relations.
Another conversion approach is \emph{Star-to-clique}, which converts a tuple defined on $k$ entities into $k \choose 2$ tuples with distinct relations between all pairwise entities in the tuple.

\begin{figure}[t]
    \centering
    \begin{subfigure}[b]{0.45\columnwidth}
        \includegraphics[width=\textwidth]{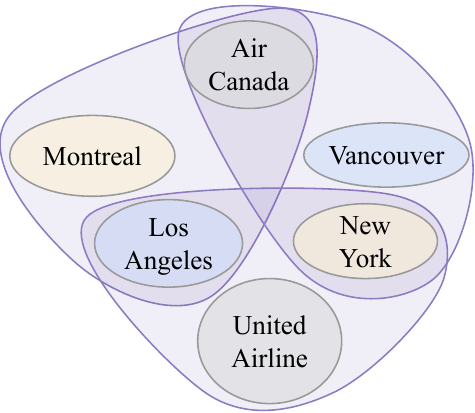}
        \caption{}
        \label{fig:hypergraph-flies}
    \end{subfigure} 
    \begin{subfigure}[b]{0.45\columnwidth}
        \includegraphics[width=\textwidth]{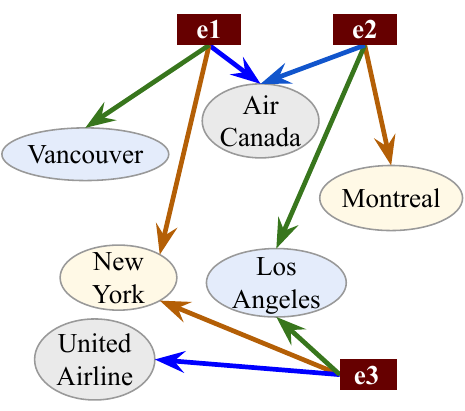}
        \caption{}
        \label{fig:reified}
    \end{subfigure} 
    \begin{subfigure}[b]{0.5\columnwidth}
        \includegraphics[width=\textwidth]{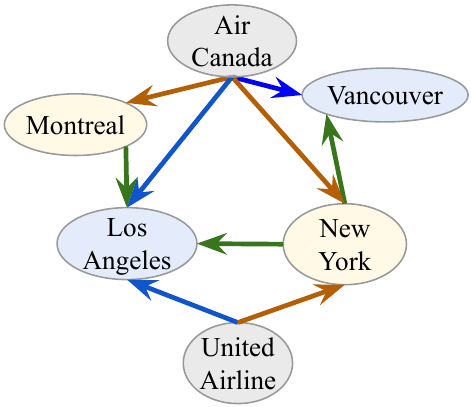}
        \caption{}
        \label{fig:star-to-clique}
    \end{subfigure}%
    \caption{
    (a) Relation \emph{flies\_between} with arity 3 defined on three tuples. (b) Reifying non-binary relations by creating three additional entities $e_1$, $e_2$, and $e_3$. (c) Converting non-binary relations into cliques using star-to-clique.
    %
    }
    \label{fig:hypergraph-to-graph}
\end{figure}

Both conversion approaches have their caveats when current link prediction models are applied to the resulting graphs. The example in Figure~\ref{fig:hypergraph-flies} shows three facts that pertain to the relation \emph{flies\_between}.
When we reify the hypergraph in this example (Figure~\ref{fig:reified}), we add three reified entities.
In terms of representation, the binary relations created are equivalent to the original representation  and reification does not lose information during conversion.
A problem with reification, however, arises at test time: because we introduce new entities that the model never encounters during training, we do not have a learned embedding for these entities; and current embedding-based methods require an embedding for each entity in order to be able to make a prediction. 

Applying the star-to-clique method to a hypergraph does not yield better results, as star-to-clique conversion loses information. Figure~\ref{fig:star-to-clique} shows the result of applying star-to-clique to the original hypergraph (Figure~\ref{fig:hypergraph-flies}), in which the tuple \textit{flies\_between(Air Canada, New York, Los Angeles)} might be interpreted as being true (since the corresponding entities are connected by edges), whereas looking at the original hypergraph, it is clear that Air Canada does not fly from New York to Los Angeles. 

In this work, we introduce two embedding-based models that perform link prediction directly on knowledge hypergraphs without converting them to graphs.
Both proposed models are based on the idea that 
predicting the existence of a relation between a set of entities depends
on the position of the entities in the relation; otherwise, the relation is symmetric. On the other hand, learning entity embeddings for each position independently does not work well either, as this does not let the information flow between the embeddings for the different positions of the same entity. 
The first model we propose is \emph{\oursimple{}}.
For a given entity, \emph{\oursimple{}} shifts the entity embedding by a value that depends on the position of the entity in the given relation.
Our second model is \emph{\ourmodel{}}, which in addition to learning entity embeddings, learns positional (convolutional) embeddings; these positional embeddings are disentangled from entity representations and are used to transform the representation of an entity based on its position in a relation. This makes \ourmodel{} more robust
to changes in the position of an entity within a tuple.
We show that both \oursimple{} and \ourmodel{} are fully expressive.
To evaluate our models, we introduce two new datasets from subsets of \acr{Freebase}, and develop baselines by extending existing models on knowledge graphs to work with hypergraphs. 


The contributions of this paper are: 
(1) showing that current techniques to convert a knowledge hypergraph to knowledge graph do not yield satisfactory results for the link prediction task,
(2) introducing \ourmodel{} and \oursimple{}, two models for knowledge hypergraph completion,
(3) a set of baselines for knowledge hypergraph completion, and
(4) two new datasets containing multi-arity relations obtained from subsets of \acr{Freebase}, which can serve as new evaluation benchmarks for knowledge hypergraph completion methods. We also show that our proposed methods outperform baselines.

\section{Related Work}

Existing methods that relate to our work in this paper can be grouped into 
the following three main categories.

\paragraph{Knowledge graph completion.} 
Embedding-based models for knowledge graph completion such as
\emph{translational}~\cite{TransE,TransH}, \emph{bilinear}~\cite{DistMult,kazemi2018simple}, and \emph{deep models}~\cite{socher2013reasoning} have proved to be effective for knowledge graphs where all relations are binary.
In Section~\ref{sec:experiments} we extend some of the models in this category to knowledge hypergraphs, and compare their performance with the proposed methods.

\paragraph{Knowledge hypergraph completion.}
Soft-rule models~\cite{de2007problog,kazemi2014relational} can easily handle variable arity relations and have the advantage of being interpretable.
However, they 
can only learn a subset of patterns~\cite{nickel2016review}.
\citeauthor{guan2019link} (\citeyear{guan2019link}) propose an embedding-based method
based on the star-to-clique approach. The caveats of this approach are discussed earlier. 
m-TransH~\cite{m-TransH} extends TransH~\cite{TransH} to knowledge hypergraph completion. \citeauthor{kazemi2018simple} (\citeyear{kazemi2018simple}) prove that TransH, and consequently m-TransH, are not fully expressive and have restrictions in modeling relations. 
In contrast, we prove that our proposed models are fully expressive.

\paragraph{Learning on hypergraphs.}
Hypergraph learning has been employed to model high-order correlations among data in many tasks, such as in video object segmentation~\cite{huang2009video} and in modeling image relationships and image ranking~\cite{huang2010image}. 
There is also a line of work extending graph neural networks to hypergraph neural networks~\cite{HGNN} and hypergraph convolution networks~\cite{HGCN}. These models are designed for undirected hypergraphs, with edges that are not labeled (no relations), while knowledge hypergraphs are directed and labeled graphs.
As there is no clear or easy way of extending these models to our knowledge hypergraph setting, we do not consider them as baselines for our experiments. 

\section{Definition and Notation}
A world consists of a finite set of entities $\entset$,
a finite set of relations $\relset$, and a set of tuples
$\tau$ where each
tuple in $\tau$ is of the form $r(e_1, e_2, \dots, e_k)$ where 
$r \in \relset$ is a relation and each $e_i\in \entset$ is an entity, for all $i=1,2,\dots,k$.
The \emph{arity} $\left|r\right|$ of a relation $r$ is the 
number of arguments that the relation takes and is fixed for each relation.
A world specifies what is true: all the tuples in $\tau$
are true, 
and the tuples that are not in $\tau$ are false. 
A knowledge hypergraph consists of a subset of the tuples $\tau' \subseteq \tau$.
Link prediction in knowledge hypergraphs is the problem of predicting the missing tuples in $\tau'$, 
that is, finding the tuples $\tau \setminus \tau'$.

An \emph{embedding} is a function that converts an entity or a relation into a vector (or sometimes a higher order tensor) over a field (typically the real numbers).
We use bold lower-case for vectors, that is, $\vc{e} \in \realset^k$ is an embedding of entity $e$, and
$\vc{r} \in \realset^{l}$ is an embedding of a relation $r$.

Let $\vc{v_1}, \vc{v_2}, \dots, \vc{v_k}$ be a set of vectors. 
The variadic function $\rm{concat}(\vc{v_1}, \dots, \vc{v_k})$ outputs the concatenation of its input vectors.
We define the variadic function $\dotsum{}$ to be the sum of the element-wise product 
of its input vectors, namely
$\dotsum{\vc{v_1}, \vc{v_2}, \dots, \vc{v_k}} \defeq  \sum_{i=1}^{\ell} \vc{v_1}^{(i)} \vc{v_2}^{(i)} \dots \vc{v_k}^{(i)}$
where each vector $\vc{v_i}$ has the same length, and $\vc{v_j}^{(i)}$ is the $i$-th element of vector $\vc{v_j}$.
The 1D convolution operator $*$ takes as input a vector $\vc{v}$, a convolution weight filter $\vc{\omega}$, and a stride $s$ and outputs the standard 1D convolution as defined in torch.nn.Conv1D function in PyTorch ~\cite{pytorch}.

For the task of knowledge graph completion, an embedding-based model defines a function $\phi$ that takes a tuple $x$
as input, and generates a prediction, \eg a probability (or score) of the tuple being true.
A model is \emph{fully expressive} if given any complete world (full assignment of truth values to all tuples), 
there exists an assignment of values to the embeddings of the entities and relations 
that accurately separates the tuples that are true in the world from those that are false.

\section{Knowledge Hypergraph Completion: Proposed Methods}
\label{HypE}
The idea at the core of our methods is that the way an entity representation is used to make predictions is affected by the role (or position) that the entity plays in a given relation. 
In the example in 
Figure~\ref{fig:hypergraph-flies}, Montreal is the departure city; but it may appear in a different position (e.g., arrival city) in another tuple.
This means that the way we use Montreal's embedding for computing predictions may need to vary 
based  on the position it appears in within the tuple. 
%
In general, when the embedding of an entity does not depend on its position in the tuple during prediction,
then the relation has to be symmetric (which is not the case for most relations). 
On the other hand, when entity embeddings are based
on position but are learned independently, information about one position will not interact with that of others.
It should be noted that in several embedding-based methods for knowledge graph completion, such as canonical polyadic~\cite{CP,lacroix2018canonical}, ComplEx~\cite{trouillon2016complex}, and SimplE~\cite{kazemi2018simple}, 
the prediction depends on the position of each entity in the tuple.

In what follows, we propose two embedding-based methods for link prediction in knowledge hypergraphs.
The first model is inspired by SimplE and has its roots in knowledge graph completion; 
the second model takes a fresh look at knowledge completion as a multi-arity problem, without first setting it up within the frame of binary relation prediction.

\begin{figure*}[t]
    \centering
    \begin{subfigure}[b]{0.68\columnwidth}
        \includegraphics[width=\textwidth]{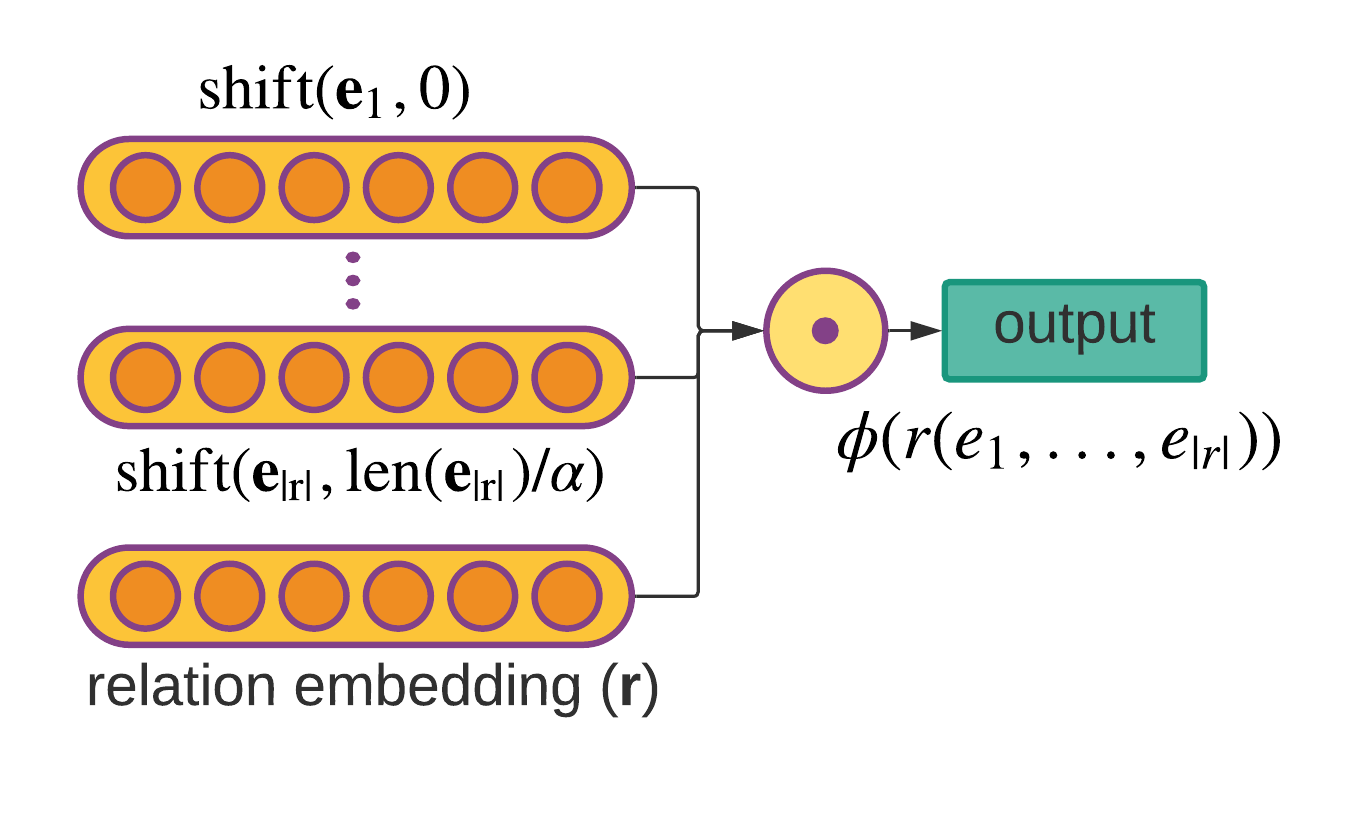}
        \caption{}
        \label{fig:score_HSimplE}
    \end{subfigure}
    \begin{subfigure}[b]{0.68\columnwidth}
        \includegraphics[width=\textwidth]{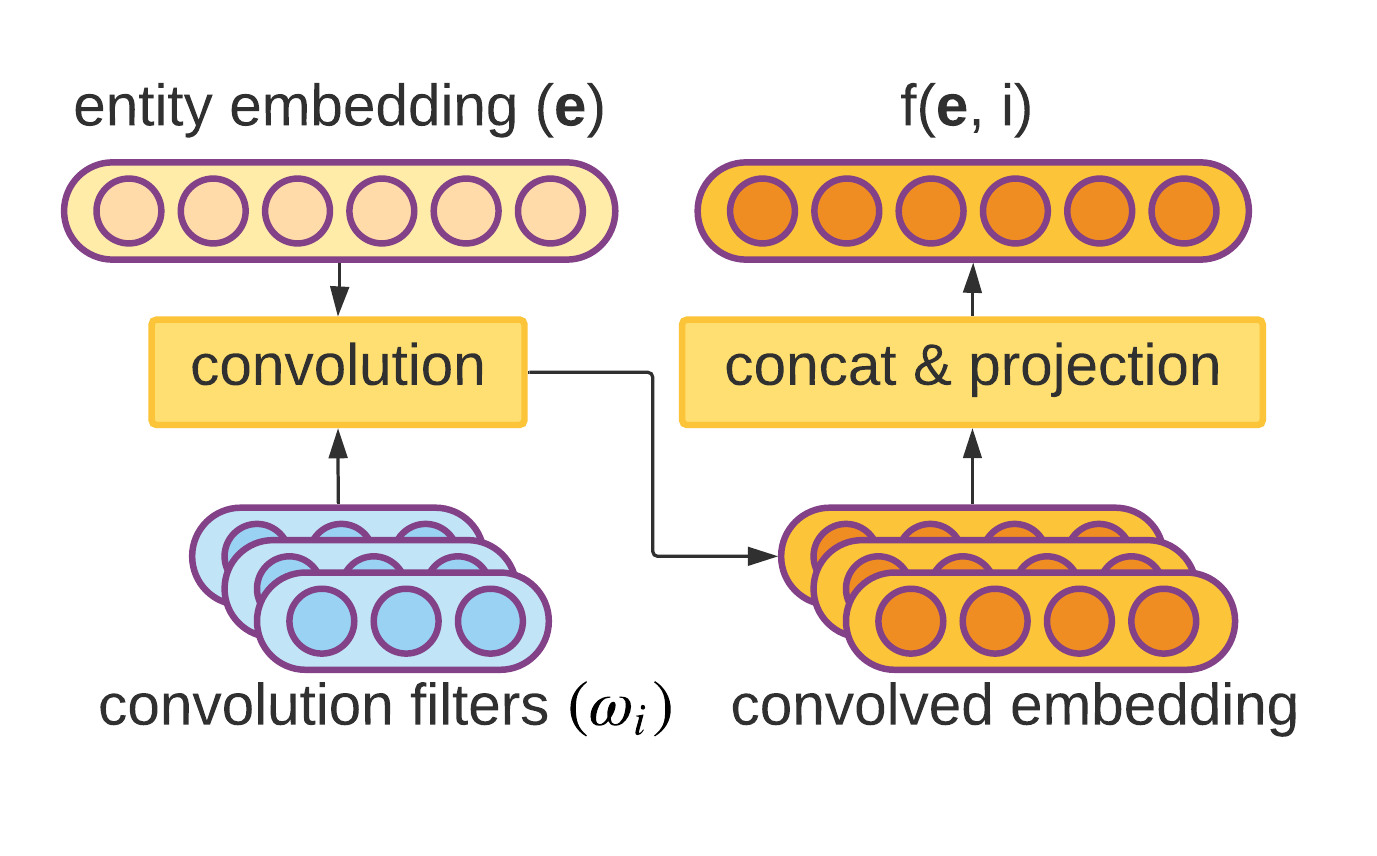}
        \caption{}
        \label{fig:representor}
    \end{subfigure}\hfill
    \begin{subfigure}[b]{0.68\columnwidth}
        \includegraphics[width=\textwidth]{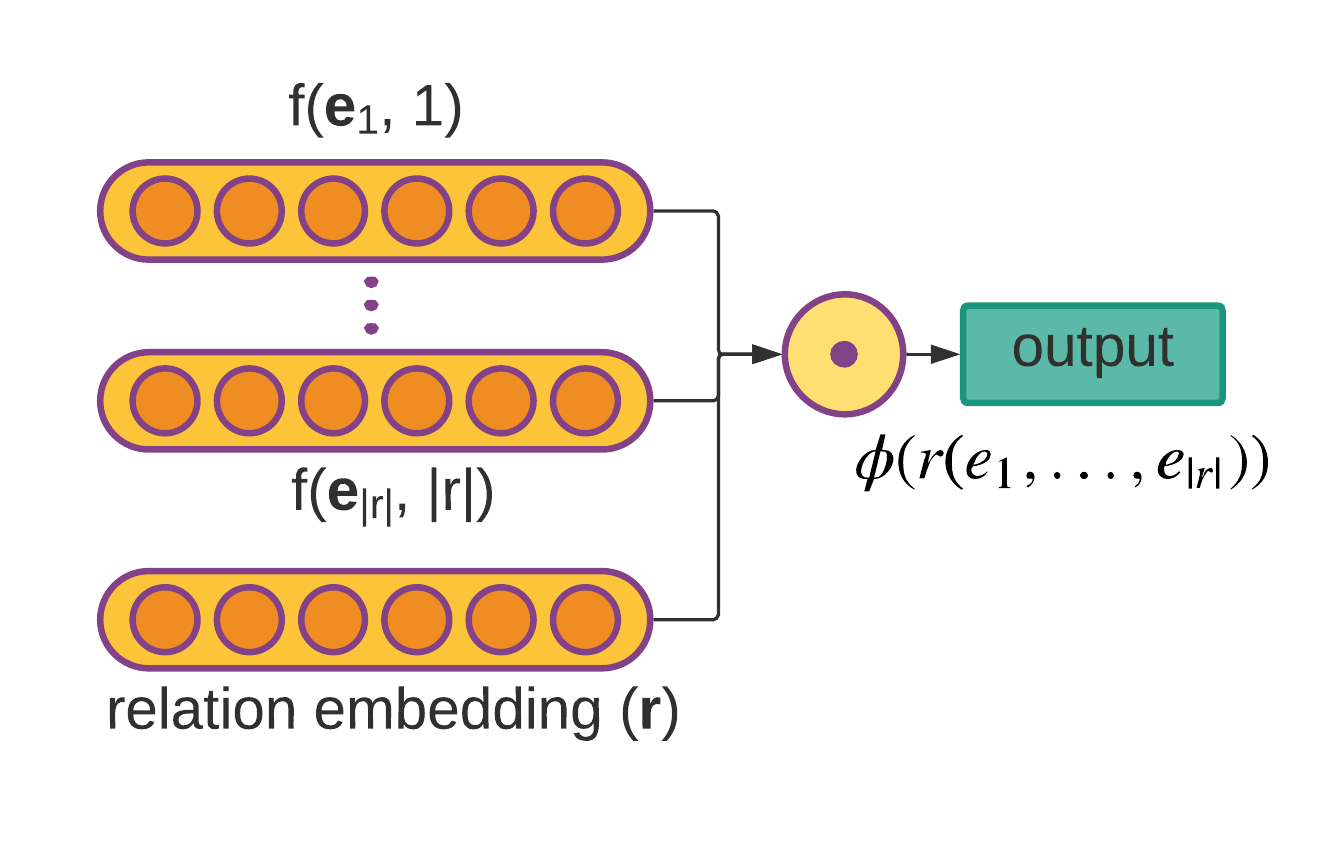}
        \caption{}
        \label{fig:score_HypE}
    \end{subfigure}
    
    \caption{Visualization of \oursimple{} and \ourmodel{} architectures. (a) $\phi$ for \oursimple{} transforms entity embeddings by shifting them based on their position and combining them with the relation embedding.
    (b) $f(\vc{e}, i)$ for \ourmodel{} takes an entity embedding and the position the entity appears in the given tuple, and returns a vector. (c) $\phi$  takes as input a tuple and outputs the score of \ourmodel{} for the tuple.}\label{fig:HyperConvE}
\end{figure*}

\paragraph{\oursimple{}.}
\oursimple{} is an embedding-based method for link prediction in knowledge hypergraphs that is inspired by SimplE~\cite{kazemi2018simple}.
SimplE learns two embedding vectors $\vc{e^{(1)}}$ and $\vc{e^{(2)}}$ for an entity~$e$
(one for each possible position of the entity),
and two embedding vectors $\vc{r^{(1)}}$ and $\vc{r^{(2)}}$ for a relation~$r$
(with one relation embedding as the inverse of the other).
It then 
computes the score of a triple as 
$\phi(r( e_1, e_2))=\dotsum{\vc{r^{(1)}},\vc{e_{1}^{(1)}},\vc{e_{2}^{(2)}}} + \dotsum{\vc{r^{(2)}},\vc{e_{2}^{(1)}},\vc{e_{1}^{(2)}}}$.
\vskip0.15cm

In \oursimple{}, we adopt the idea of having different representations for an entity
based on its position in a relation and updating all these representations
from a single training tuple. 
We do this by representing each entity $e$ as a single vector
$\vc{e}$ (instead of multiple vectors as in SimplE), and 
each relation $r$ as a single vector $\vc{r}$.
Conceptually, each $\vc{e}$ can be seen as the concatenation of the different representations
of $e$ based on every possible position.
For example, in a knowledge hypergraph where the relation with maximum arity is $\maxar{}$, 
an entity can appear in $\maxar{}$ different positions; hence $\vc{e}$ will be the concatenation 
of $\maxar{}$ vectors, one for each possible position.
\oursimple{} scores a tuple using the following function.
\begin{multline}
\phi(r(e_i, e_j, \dots, e_k)) = \dotsum{\vc{r}, \vc{e_i}, \rm{shift}(\vc{e_j}, \rm{len}(\vc{e_j})/\maxar{}), \dots,\\ \rm{shift}(\vc{e_k}, \rm{len}(\vc{e_k})\cdot(\maxar{}-1)/\maxar{}))}
\end{multline}
Here, $\rm{shift}(\vc{v},$ $x)$ 
shifts vector $\vc{v}$ to the left by $x$ steps, $\rm{len}(\vc{e})$ returns length of vector $\vc{e}$, and
$\maxar{} = \max_{r \in \relset}(|r|)$.
We observe that for knowledge graphs ($\maxar{}=2$), 
SimplE is a special instance of \oursimple{}, with 
$\vc{e} = \rm{concat}(\vc{e}^{(1)}, \vc{e}^{(2)})$ and $\vc{r}= \rm{concat}(\vc{r}^{(1)}, \vc{r}^{(2)})$. 
The architecture of \oursimple{} is summarized in Figure~\ref{fig:score_HSimplE}.

\paragraph{\ourmodel{}.}
\ourmodel{} learns a single representation for each entity, a single representation
for each relation, and positional convolutional weight filters for each possible position.
When an entity appears in a specific position, the appropriate positional filters are first used to transform the embedding 
of each entity in the given fact; 
these transformed entity embeddings 
are then combined with the embedding of the relation to produce a~\emph{score}, 
\ie the probability of the input tuple to be true. 
The architecture of \ourmodel{} is summarized in Figures~\ref{fig:representor} and \ref{fig:score_HypE}.

Let $n$ denote the number of filters per position, $l$ the filter-length, $d$ the embedding dimension, and $s$ the stride of a convolution.
Let $\omega_i \in \mathbb{R}^{n \times l}$ be the convolutional 
filters associated with position $i$, and let 
$\omega_{ij} \in \mathbb{R}^l$ be the $j$th row of $\omega_i$.
We denote by $P \in \mathbb{R}^{n q \times d}$ the projection matrix,
where $q = \floor{(d - l) / s} + 1$ is the feature map size. 
For a given tuple, define $f(\vc{e}, i) = \rm{concat}(\vc{e} * \omega_{i1}, \dots, \vc{e} * \omega_{i n})P$
to be a function that returns a vector of size $d$ based on the entity embedding $\vc{e}$ and its position $i$ in the tuple. 
Thus, each entity embedding $\vc{e}$ appearing at position $i$ in a given tuple is convolved with the set of position-specific filters $\omega_{i}$ to give $n$ feature maps of size~$q$.
All $n$ feature maps corresponding to an entity are concatenated into a vector
of size $nq$ and projected to the embedding space through multiplication by $P$.
The projected vectors of entities and the embedding of the relation 
are combined by inner-product to define $\phi$:
\begin{equation}
\label{eq:phi}
    \phi(
    r(e_1, \dots, e_{|r|})) = \tuplee{\vc{r}, f(\vc{e_1}, 1), \dots, f(\vc{e_{|r|}}, |r|)}.
\end{equation}

The advantage of learning positional filters disentangled from entity embeddings is two-folds:
On one hand, learning a single vector per entity keeps entity representations simple and disentangled from its position in a given fact.
On the other hand, unlike \oursimple{}, \ourmodel{} learns positional filters from all entities that appear in the given position; 
overall, this separation of representations for entities, relations, and positions facilitates the representation of knowledge bases having facts of arbitrary number of entities. It also gives \ourmodel{} an additional robustness, such as in the case when we test a trained \ourmodel{} model on a tuple that contains an entity in a position never seen before at train time. We discuss this 
in Section~\ref{sec:hypergraph results}.

Both \oursimple{} and \ourmodel{} are fully expressive --- an important property that has been 
the focus of several studies~\cite{simpleplus}. 
A model that is not fully expressive can 
embed assumptions that may not be reflected in reality. 

\begin{theorem}[Expressivity]\label{theorem:expressivity}
For any ground truth over entities \entset{} and relations \relset{} containing $|\tau|$ true tuples and ${\maxar{} = \max_{r \in \relset}(|r|)}$
, there exists a \ourmodel{} and a \oursimple{} model with embedding vectors of size $\max(\maxar{} |\tau|, \maxar{})$ that represents that ground truth.
\end{theorem}
\begin{hproof}
To prove the theorem, we show an assignment of embedding values for each of the entities and relations in $\tau$ such that the scoring function of \ourmodel{} and \oursimple{} gives~$1$ for $t \in \tau$ and $0$ otherwise. 
\end{hproof}
\subsection{Objective Function and Training}
Both \oursimple{} and \ourmodel{} are trained using stochastic gradient descent with mini-batches.
In each learning iteration, we take a batch of positive tuples from the knowledge hypergraph. As we only have positive instances available, we need to also train our model on negative instances; thus, for each positive instance, we produce a set of negative instances. 
For negative sample generation, we follow 
the contrastive approach of \cite{TransE} for knowledge graphs and extend it to knowledge hypergraphs: 
for each tuple, 
we produce a set
of negative samples of size $N |r|$ 
by replacing each of the entities with $N$ random entities in the tuple, one at a time. 
Here, $N$ is the ratio of negative samples in our training set and is a hyperparameter.

Given a knowledge hypergraph defined on $\tau'$, we let $\tau'_{train}$, $\tau'_{test}$, and $\tau'_{valid}$ denote
the train, test, and validation sets, respectively,
so that $\tau' = \tau'_{train} \cup \tau'_{test}\cup \tau'_{valid}$. 
For any tuple $x$ in $\tau'$, we let $T_{neg}(x)$ be a function that generates a 
set of negative samples through the process described above.
Let $\vc{r}$ and $\entityv$ represent relation and entity embeddings respectively.
%
We define the following cross entropy loss:
\begin{equation*}
\mathcal{L}(\vc{r}, \entityv) =  \sum_{x' \in \tau'_{train}}{-log\bigg(\frac{e^{\phi(x')}}{ e^{\phi(x')} + \displaystyle\sum_{x \in T_{neg}(x')}{e^{\phi(x)}}}\bigg)}
\end{equation*}

\section{Experimental Setup}

\begin{table*}
\footnotesize
\setlength{\tabcolsep}{2pt}

\begin{center}
\setlength{\tabcolsep}{3pt}
\begin{tabular}{l|cccc|cccc|cccc}
\multicolumn{1}{c}{}
& \multicolumn{4}{c}{JF17K}
& \multicolumn{4}{c}{\acr{FB-auto}} & \multicolumn{4}{c}{\acr{m-FB15K}}\\
\cmidrule(lr){2-5} \cmidrule(lr){6-9}
\cmidrule(lr){10-13}
Model    & MRR   & Hit@1 & Hit@3 & Hit@10 & MRR & Hit@1 & Hit@3 & Hit@10 & MRR & Hit@1 & Hit@3 & Hit@10\\\hline
r-SimplE & 0.102 & 0.069 & 0.112 & 0.168 & 0.106 & 0.082 & 0.115 & 0.147 & 0.051 & 0.042 & 0.054 & 0.070\\
m-DistMult & 0.463 & 0.372 & 0.510 & 0.634 & 0.784 & 0.745 & 0.815 & 0.845 & 0.705 & 0.633 & 0.740 & 0.844 \\
m-CP & 0.391 & 0.298 & 0.443 & 0.563 & 0.752& 0.704 & 0.785 & 0.837 & 0.680 & 0.605 & 0.715 & 0.828 \\
m-TransH
~\cite{m-TransH} & 0.444 & 0.370 & 0.475 & 0.581 & 0.728 & 0.727 & 0.728 & 0.728& 0.623 & 0.531 & 0.669 & 0.809\\
\hline
\oursimple{} (Ours) & 0.472 & 0.378 & 0.520 & 0.645 & 0.798 & 0.766 & 0.821 & 0.855 & 0.730 & 0.664 & 0.763 & 0.859\\
\ourmodel{} (Ours) & \textbf{0.494} & \textbf{0.408} & \textbf{0.538} & \textbf{0.656} & \textbf{0.804} &
 \textbf{0.774} &  \textbf{0.823} & \textbf{0.856} & \textbf{0.777} & \textbf{0.725} & \textbf{0.800} & \textbf{0.881}  \\
\end{tabular}
\end{center}
\caption{Knowledge hypergraph completion results on JF17K, \acr{FB-auto} and \acr{m-FB15K} for baselines and the proposed method. The prefixes `r' and `m' in the model names stand for \emph{reification} and \emph{multi-arity} respectively. Both our methods outperform the baselines on all datasets. 
}
\label{table:FB-subsets}
\end{table*}

\subsection{Datasets}
The experiments on knowledge hypergraph completion are conducted on three datasets. The first is 
JF17K proposed by \citeauthor{m-TransH} (\citeyear{m-TransH}); as no validation set is proposed
for JF17K, we randomly select 20\% of the train set as validation.
We also create two datasets \acr{FB-auto} and \acr{m-FB15K} from \acr{Freebase}. 
For the experiments on datasets with binary relations, we use two standard benchmarks for 
knowledge graph completion: WN18~\cite{WN18} and FB15k~\cite{TransE}. 
See Table~\ref{statistics} for statistics of the datasets.

\begin{table}
\footnotesize
\centering
\setlength{\tabcolsep}{3pt}
\begin{tabular}{@{} l| *5c @{}}
\multicolumn{1}{c|}{Dataset}& $|\mathcal{E}|$  & $|\mathcal{R}|$  & \#train  & \#valid & \#test\\ 
\hline
 WN18 & 40,943 & 18 & 141,442 & 5,000& 5,000\\ 
 FB15k & 14,951 & 1,345 & 483,142 & 50,000 & 59,071\\ \hline
 JF17K & 29,177 & 327 & 77,733 & -- & 24,915\\
\acr{FB-auto} & 3,410 & 8 & 6,778 & 2,255 & 2,180\\
\acr{m-FB15K} & 10,314 & 71 & 415,375 & 39,348 & 38,797 \\
\end{tabular}
\caption{Dataset Statistics.}\label{statistics}
\end{table}

\subsection{Baselines}\label{baselines}
To compare our results to that of existing work, we first design simple baselines that extend current models to work with knowledge hypergraphs.
We only consider models that admit a simple extension to higher-binary relations for the link prediction task.
The baselines for this task are grouped into the following categories:
(1) methods that work with binary relations and that are easily extendable to higher-arity, namely r-SimplE, m-DistMult, and m-CP;
(2) existing methods that can handle higher-arity relations, namely m-TransH. 
Below we give some details about methods in category (1).

\paragraph{r-SimplE.} To test the performance of a model trained on reified data,
we convert higher-arity relations in the train set to binary relations through reification. We then use SimplE (that we call r-SimplE) on this reified data. In this setting, at test time higher-arity relations are first reified to a set of binary relations; this process creates new auxiliary entities for which the model has no learned embeddings. To embed the auxiliary entities for the prediction step,
we use the observation we have about them at test time.
For example, a higher-arity relation $r(e_1, e_2, e_3)$ is reified at test time by adding a new entity $e'$ and converting the higher-arity tuple to three binary facts: $r_1(e', e_1)$, $r_2(e', e_2)$, and $r_3(e', e_3)$. 
When predicting the tail entity of $r_1(e', ?)$, we use the other two reified facts to 
learn an embedding for entity~$e'$. 
Because $e'$ is added only to help represent the higher-arity relations as a set of binary relations, 
we only need to do tail prediction for reified relations. Note that we do not reify binary relations.

\paragraph{m-DistMult.} DistMult~\cite{DistMult} defines a score function $\phi(r(e_i, e_j)) = \dotsum{\vc{r}, \vc{e_i}, \vc{e_j}}$. To accommodate non-binary relations, we redefine this function as 
$\phi(\tupleexample) = \dotsum{\vc{r}, \vc{e_i}, \dots, \vc{e_j}}$. 

\paragraph{m-CP.} Canonical Polyadic decomposition~\cite{CP} is a tensor decomposition approach. We refer to the version that only handles binary relations as \emph{CP}. CP embeds each entity $e$ as two vectors $\vc{e^{(1)}}$ and $\vc{e^{(2)}}$, and each relation $r$ as a single vector $\vc{r}$;
it defines the score function ${\phi(r(e_i, e_j))=\dotsum{\vc{r},\vc{e_{i}^{(1)}},\vc{e_{j}^{(2)}}}}$. 
We extend CP to \emph{m-CP}, which accommodates 
relations of any arity. m-CP embeds each entity $e$ as $\maxar{}$ different vectors $\vc{e^{(1)}}, .., \vc{e^{(\maxar{})}}$, where $\maxar{} = \max_{r \in \relset}(|r|)$; it computes the score of a tuple as $\phi(\tupleexample)=\dotsum{\vc{r}, \vc{e_i^{(1)}}, ..., \vc{e_j^{(|r|)}}}$.

\subsection{Evaluation Metrics}
Given a knowledge hypergraph on $\tau'$, we evaluate various completion methods using a train and test set $\tau'_{train}$ and $\tau'_{test}$.
We use two evaluation metrics: Hit@t and Mean Reciprocal Rank~(MRR). 
Both these measures rely on the \emph{ranking} of a tuple $x \in \tau'_{test}$ within a set of \emph{corrupted} tuples.
For each tuple  $r(e_1, \dots, e_k)$ in $\tau'_{test}$ and each entity position $i$ in the tuple,
we generate $|\entset|-1$ corrupted tuples by replacing 
the entity $e_i$ with each of the entities in $\entset \setminus \{e_i\}$.
For example, by corrupting entity $e_i$, we would obtain a new tuple $r(e_1, \dots, e_i^c, \dots, e_k)$ where $e_i^c \in \entset \setminus \{e_i\}$.
Let the set of corrupted tuples, plus $r(e_1, \dots, e_k)$, be denoted by $\theta_i(r(e_1,\dots,e_k))$.
Let $\rank_i(r(e_1, \dots, e_k))$ be the ranking of $r(e_1, \dots, e_k)$ within 
$\theta_i(r(e_1, \dots, e_k))$ based on the score $\phi(x)$ for each $x \in \theta_i(r(e_1,\dots,e_k))$.
We compute the MRR as
$
\frac{1}{K} \sum_{r(e_1, \dots,e_k) \in \tau'_{test}} 
\sum_{i=1}^{k}\frac{1}{\rank_{i} r(e_1, \dots,e_k)}
$
where $K = \sum_{r(e_1,\dots e_k) \in \tau'_{test}} |r|$ is the number of prediction tasks. 
Hit@t measures the proportion of tuples in $\tau'_{test}$ that rank among the top $t$ in their corresponding corrupted sets. 
We follow \cite{TransE} and remove all corrupted tuples that are in $\tau'$ from our computation of MRR and Hit@t.

\begin{table*}
\footnotesize
\begin{center}
\setlength{\tabcolsep}{3pt}
\begin{tabular}{l|cccc|cccc}
\multicolumn{1}{c}{}
& \multicolumn{4}{c}{WN18} & \multicolumn{4}{c}{FB15k}\\
\cmidrule(lr){2-5} \cmidrule(lr){6-9}
Model    & MRR   & Hit@1 & Hit@3 & Hit@10 & MRR & Hit@1 & Hit@3 & Hit@10\\\hline
CP~\cite{CP} & 0.074 & 0.049 & 0.080 & 0.125 & 0.326 & 0.219 & 0.376 & 0.532  \\
TransH~\cite{TransH} & - & - & - & 0.867 & - & - & - & 0.585\\
m-TransH~\cite{m-TransH} & 0.671 & 0.495 & 0.839 & 0.923 & 0.351 & 0.228 & 0.427 & 0.559 \\
DistMult~\cite{DistMult} & 0.822 & 0.728 & 0.914 & 0.936  & 0.654 & 0.546 & 0.733 & 0.824\\
HSimplE (Ours) and SimplE~\cite{kazemi2018simple}   & \textbf{0.942} & \textbf{0.939} & \textbf{0.944} & \textbf{0.947}  & \textbf{0.727} & \textbf{0.660} & 0.773 & 0.838\\
\ourmodel{} (Ours) & 0.934 & 0.927 & 0.940 & 0.944 & 0.725 & 0.648 & \textbf{0.777} & \textbf{0.856}\\
\end{tabular}
\caption{Knowledge graph completion results on WN18 and FB15K for baselines and \ourmodel{}. Note that in knowledge graphs (binary relations), \oursimple{} and SimplE are equivalent, both theoretically and experimentally. The results show that our methods outperform the baselines.
}
\label{table:results-table-arity2}
\end{center}
\end{table*}


\begin{table}
\footnotesize
\begin{center}
\setlength{\tabcolsep}{3pt}
\begin{tabular}{l|ccc|c}
\multicolumn{1}{c}{}
& \multicolumn{3}{c}{Arity}\\
\cmidrule(lr){2-4}
Model    & 2 & 3 & 
4-5-6 & All\\\hline
r-SimplE & 0.478 & 0.025 &
0.017 & 0.168\\
m-DistMult & 0.495 & 0.648 & 
0.809 & 0.634\\
m-CP & 0.409 & 0.563 &
0.765 & 0.560\\
m-TransH~\cite{m-TransH} & 0.411 &	0.617 & 
0.826 & 0.596\\
\hline
\oursimple{} (Ours) & \textbf{0.497} & \textbf{0.699} & 
0.745 & 0.645\\
\ourmodel{} (Ours) & 0.466	& 0.693 & 
\textbf{0.858} & \textbf{0.656}\\
\hline\hline
\# train tuples & 36,293 & 18,846 & 
6,772 & 61,911\\
\# test tuples & 10,758 & 10,736 & 
3,421 & 24,915\\
\end{tabular}
\end{center}
\caption{Breakdown performance of Hit@10 across relations with different arities on JF17K dataset along with their statistics. 
}
\label{table:ablation-study}
\end{table}

\begin{figure}
     \centering
     \begin{subfigure}[b]{0.70\columnwidth}
         \centering
         \includegraphics[width=\textwidth]{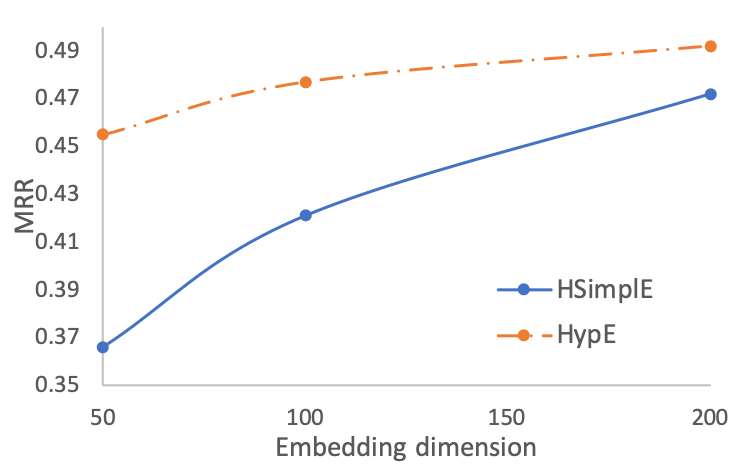}
         \caption{}
         \label{fig:constrained_budget}
     \end{subfigure}
     \hfill
     \begin{subfigure}[b]{0.7\columnwidth}
         \centering
         \includegraphics[width=\textwidth]{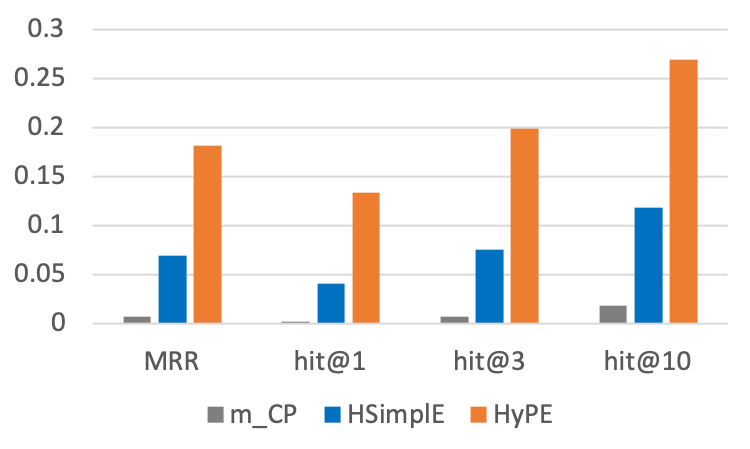}
         \caption{}
         \label{fig:difficult}
     \end{subfigure}
        \caption{
        These experiments show that \ourmodel{} outperforms \oursimple{} when trained with fewer parameters, and when tested on samples that contain at least one entity in a position never encountered during training.
        (a) MRR of \ourmodel{} and \oursimple{} for different embedding dimensions. (b) Results of m-CP, \oursimple{}, and \ourmodel{} on the \emph{missing positions} test set (containing $1{,}806$ test samples).}
        \label{fig:Plots}
\end{figure}

\section{Experiments}\label{sec:experiments}
This section summarizes our experiments
\footnote{Code and data available at \href{https://github.com/ElementAI/HypE}{https://github.com/ElementAI/HypE}.}.

\subsection{Knowledge Hypergraph Completion Results}
\label{sec:hypergraph results}
The results of our experiments, summarized in Table~\ref{table:FB-subsets}, show that
both 
\oursimple{} and \ourmodel{} outperform the proposed baselines across the three datasets JF17K, \acr{FB-auto}, and \mbox{\acr{m-FB15K}}. 
They further demonstrate that reification for the \mbox{r-SimplE} model does not work well; this is because the reification process introduces auxiliary entities for which the model does not learn appropriate embeddings because these auxiliary entities appear in very few facts. 
Comparing the results of r-SimplE against \oursimple{}, we can also see that extending a model to work with hypergraphs works better than reification when high-arity relations are present.

The ability of knowledge sharing through the learned position-dependent convolution filters suggests that \ourmodel{} would need a lower number of parameters than \oursimple{} to obtain good results. To test this, we train both models with different embedding dimensions.
Figure~\ref{fig:constrained_budget} shows the MRR on the test set for each model with different embedding sizes. Based on the MRR result, we can see that \ourmodel{} outperforms \oursimple{} by $24\%$ for embedding dimension $50$, implying that \ourmodel{} works better under a constrained budget. 

Disentangling the representations of entity embeddings and positional filters enables \ourmodel{} to better learn the role of position within a relation because the learning process considers the behavior of all entities that appear in a given position at the time of training. This becomes especially important in the case when some entities never appear in certain positions in the train set, but you still want to be able to reason about them no matter what position they appear in at test time.
In order to test the effectiveness of our models in this more challenging scenario, we created a \emph{missing positions} test set by selecting the tuples from our original test set
that contain at least one entity in a position it never appears in within the training dataset.
The results on these experiments (Figure~\ref{fig:difficult}) show that (1) both \oursimple{} and \ourmodel{} outperform m-CP (which learns different embeddings for each \mbox{entity-position} pair), and more importantly, (2) \ourmodel{} significantly outperforms \oursimple{} for this challenging test set, leading us to believe that disentangling entity and position representations may be a better strategy for this scenario.

\subsection{Knowledge Graph Completion Results}
To show that \oursimple{} and \ourmodel{} work well also on the more common knowledge graphs, we evaluate them on WN18 and FB15K. 
Table~\ref{table:results-table-arity2} shows link prediction results on WN18 and FB15K. 
Baseline results are taken from the original papers except that of m-TransH, which we implement ourselves.
Instead of tuning the parameters of \ourmodel{} to get potentially better results, we follow the \citeauthor{kazemi2018simple} (\citeyear{kazemi2018simple}) setup with the same grid search approach by setting $n = 2$, $l = 2$, and $s = 2$.
This results in all models in Table~\ref{table:results-table-arity2} having the same number of parameters, and thus makes them directly comparable to each other.
Note that for knowledge graph completion (all binary relations) \oursimple{} is equivalent to SimplE, both theoretically and experimentally (as shown in Section~\ref{HypE}). 
The results show that on WN18 and FB15K, \oursimple{} and \ourmodel{} outperform all baselines.

\subsection{Ablation Study on Different Arities}
We break down the performance of models across different arities. As the number of test tuples in higher arities (4-5-6) is much less than in smaller arities (2-3), we used equivalent size bins to show the decomposed results for a reasonable number of test tuples. Table~\ref{table:ablation-study} shows the Hit@10 results of the models for bins of arity 2, 3, and 4-5-6 in JF17K. 
The proposed models outperform the state-of-the-art and the baselines in all arities.
We highlight that r-SimplE and \oursimple{} are quite different models for relations having arity $> 2$.

\section{Conclusions}
Knowledge hypergraph completion is an important problem that has received little attention.
Having introduced two new knowledge hypergraph datasets, baselines, and two new methods for link prediction in knowledge hypergraphs, we hope to kindle interest in the problem. Unlike graphs, hypergraphs have a more complex structure that opens the door to more challenging questions such as: how do we effectively predict the missing entities in a given (partial) tuple?

\bibliographystyle{named}

\section*{Supplementary Material}

\begin{table*}[t]
\footnotesize
\setlength{\tabcolsep}{2pt}
\vspace{-3mm}
\begin{center}
\setlength{\tabcolsep}{3pt}
\begin{tabular}{l|ccccc|ccccccc}
\multicolumn{3}{c}{}
& \multicolumn{3}{c}{number of tuples}
& \multicolumn{5}{c}{number of tuples with respective arity}\\
\cmidrule(lr){4-6} \cmidrule(lr){7-11}
Dataset    & $|\mathcal{E}|$  & $|\mathcal{R}|$  & \#train  & \#valid & \#test & \#arity=2 & \#arity=3& \#arity=4& \#arity=5& \#arity=6 \\ \hline
 WN18 & 40,943 & 18 & 141,442 & 5,000& 5,000 & 151,442 & 0 & 0 & 0 & 0\\ 
 FB15k & 14,951 & 1,345 & 483,142 & 50,000 & 59,071 & 592,213 & 0 & 0 & 0 & 0 \\ \hline
 JF17K & 29,177 & 327 & 77,733 & -- & 24,915 & 56,322 & 34,550 & 9,509 & 2,230 & 37\\
\acr{FB-auto} & 3,410 & 8 & 6,778 & 2,255 & 2,180 &  3,786 & 0 & 215 & 7,212 & 0\\
\acr{m-FB15K} & 10,314 & 71 & 415,375 & 39,348 & 38,797 & 82,247 & 400,027 & 26 & 11,220& 0\\
\end{tabular}
\end{center}
\caption{Dataset Statistics.}\label{statistics-sup}
\end{table*}

\subsection*{Dataset Creation}

We downloaded \acr{Freebase}~\cite{bollacker2008freebase} from \url{https://developers.google.com/freebase}. 
Note  that \acr{Freebase} is a reified dataset; 
that is, it is created from a knowledge base having
facts with relations defined on two or more entities. Relation names in \acr{Freebase} have three parts separated with a `.' e.g., \texttt{film.performance.actor}. The first part in the relation shows the subject of this triple, the second part shows the actual relation, and the third part shows the role in the relation.

Here are three examples of binary relations in \acr{Freebase} with the subject of "film".
\begin{itemize}
\item \texttt{film.performance.actor(/m/04q4bd8, /m/07yyhx)}
\item \texttt{film.performance.character(/m/04q4bd8, /m/0qzpnz5)} 
\item \texttt{film.performance.film(/m/04q4bd8, /m/04mxv5p)}
\end{itemize}
From the 
{Knowledge Graph Search API Link}, (\url{https://developers.google.com/knowledge-graph/reference/rest/v1/?apix=true&apix_params=\%7B\%22ids\%22\%3A\%5B\%22\%2Fm\%2F04mxv5p\%22\%5D\%7D}) the entities \texttt{/m/07yyhx}, \texttt{/m/0qzpnz5}, and \texttt{/m/04mxv5p} refer, respectively, to actress ``Kathlyn Williams'', character ``Queen Isabel of Bourbon'', and movie ``The Spanish Dancer''.  
The entity \texttt{/m/04q4bd8} is a reified entity. 

To obtain a knowledge hypergraph $H$ from \acr{Freebase}, we perform an inverse reification process
by following the steps below.

\begin{enumerate}
\item From \acr{Freebase}, remove the facts that have relations defined on a single entity, or that contain numbers or enumeration as entities. 
\item Join the triples in \acr{Freebase} that have the same subject and actual relation, and the same entity. The role (the 3rd part of the relation) provides the position is the resulting relation.

For example, we convert the three tuples above to a tuple of relation of arity $3$ as:
\begin{center}
\texttt{film.performance(/m/07yyhx, /m/0qzpnz5, /m/04mxv5p)} 
\end{center}
with the first, 
second, and third positions representing the actor, character, and movie of the performance respectively. 
in $H$.
\item Create the \acr{FB-auto} dataset by selecting the facts from $H$ whose subject is `automotive' (first part of relation names is `automative') .
\item Create the \acr{m-FB15K} dataset by following a strategy similar to that proposed by \cite{TransE}: select the facts in $H$ that pertain to entities present in the Wikilinks database~\cite{wikilinks}.
\item Split the facts in each of \acr{FB-auto} and \acr{m-FB15K} randomly into train, test, and validation sets.
\end{enumerate}

\subsection*{Database Statistics}

Table~\ref{statistics-sup} summarizes the statistics of the datasets.

\begin{figure*}[t]
    \centering
    \includegraphics[width=0.6\textwidth]{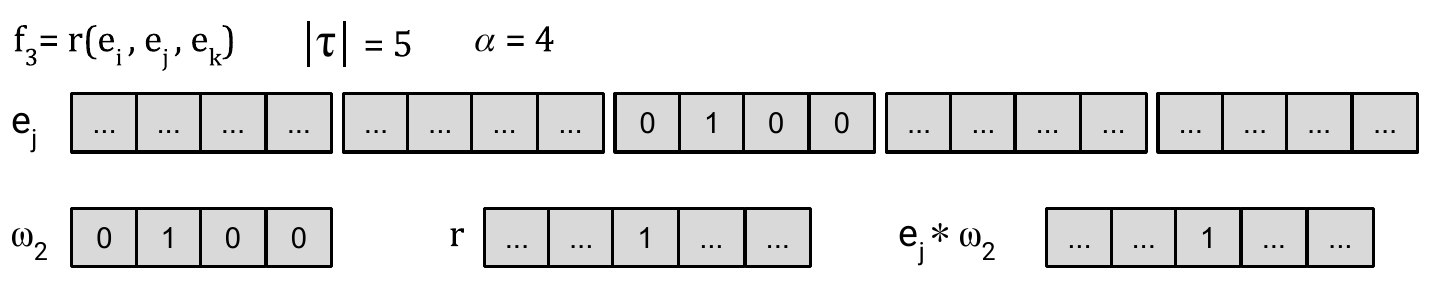}
        \caption{An example of an embedding where $|\tau| = 5$, $\maxar=4$ and $f_3$ is the third fact in $\tau$}
        \label{fig:expressivity}
\end{figure*}

\subsection*{Theorems}

\begin{theorem}[Expressivity of \ourmodel{}]\label{theorem:expressivity-sup1}
Let $\tau$ be a set of true tuples defined over entities $\entset$ and relations $\relset$, and let
$\maxar = \max_{r \in \relset}(|r|)$ be the maximum arity of the relations in $\relset$.
There exists a \ourmodel{} model with embedding vectors of size $\max(\maxar |\tau|, \maxar)$ 
that assigns $1$ to
the tuples in $\tau$ and $0$ to others.
\end{theorem}

\begin{figure*}[b]
    \centering
    \includegraphics[width=0.6\textwidth]{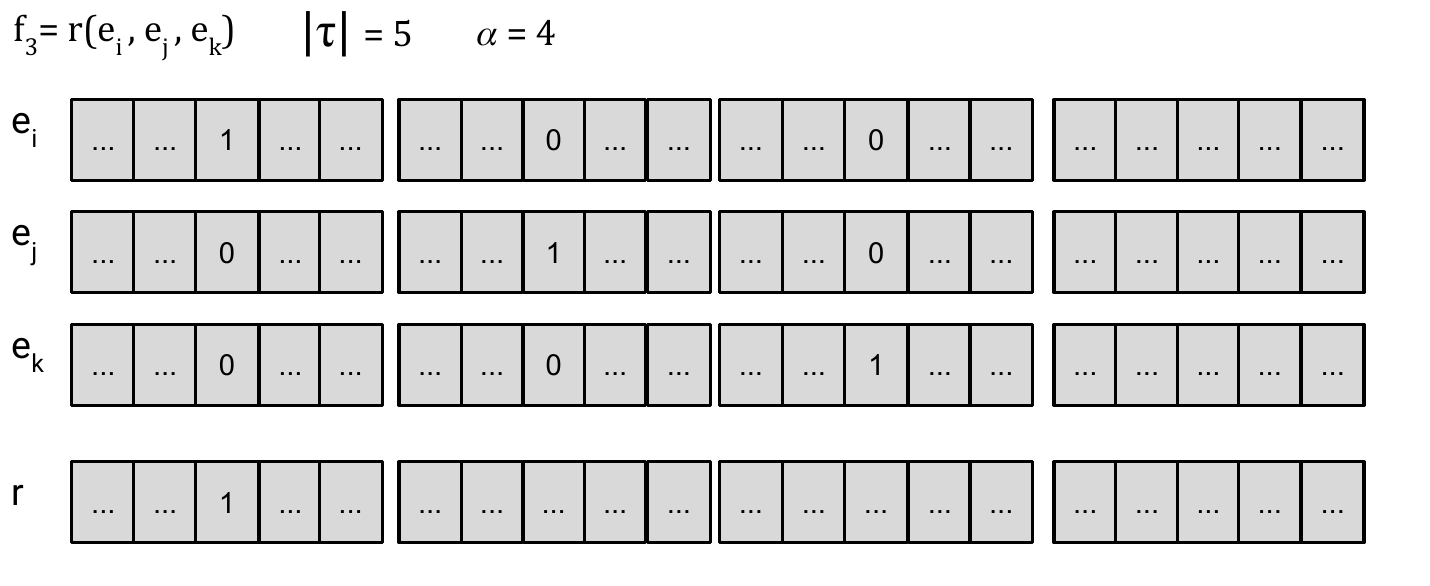}
        \caption{An example of an embedding where $|\tau| = 5$, $\maxar=4$ and $f_3$ is the third fact in $\tau$}
        \label{fig:expressivity2}
\end{figure*}

\begin{proof}
To prove the theorem, we show an assignment of embedding values for each entity and relation
in $\tau$ such that the scoring function of \ourmodel{} is as follows:
\[
  \phi(x) =
  \begin{cases}
                                  1 & \text{if $x \in \tau$} \\
                                  0 & \text{otherwise}
  \end{cases}
\]

We begin the proof by first describing the embeddings of each of the entities and relations in \ourmodel{};
we then proceed to show that with such an embedding, \ourmodel{} can represent any world accurately.

Let us first assume that $|\tau| > 0$ and let $f_p$ be the $p$th fact in $\tau$.
We let each entity $e \in \entset$ be represented with a vector of length $\maxar |\tau|$ in which  
the $p$th block of $\maxar$-bits is the one-hot representation of $e$ in fact $f_p$: if $e$ appears in fact $f_p$ at position $i$, 
then the $i$th bit of the $p$th block is set to $1$, and to $0$ otherwise. 
Each relation $r \in \relset$ is then represented as a vector of length $\tau$ whose $p$th bit 
is equal to $1$ if fact $f_p$ is defined on relation $r$, and $0$ otherwise. 

\ourmodel{} defines convolutional weight filters for each entity position within a tuple.
As we have at most $\maxar$ possible positions, we define each convolutional filter
$\omega_i$ as a vector of length $\maxar$ where the $i$th bit is set to $1$ and all others to $0$, for each $i=1,2,\dots,\maxar$.
When the scoring function $\phi$ is applied to some tuple $x$, for each entity position $i$ in $x$,
convolution filter $\omega_i$ is applied to the entity at position $i$ in the tuple as a first step; 
the $\dotsum{}$ function is then applied to the resulting vector and the relation embedding to obtain a score.

Given any tuple $x$, we want to show that $\phi(x) = 1$ if $x \in \tau$ and $0$ otherwise.

First assume that $x=f_p$ is the $p$th fact in $\tau$
that is defined on relation $r$ and entities where $e_i$ is the entity at position $i$.
Convolving each $e_i$ with $\omega_i$ results in 
a vector of length $|\tau|$ where the $p$th bit is equal to $1$
(since both $\omega_i$ and the $p$th block of $e_i$ have a $1$ at the $i$th position)
(See Figure~\ref{fig:expressivity}.
Then, as a first step, function \dotsum{} computes the element-wise multiplication between the embedding of relation $r$ 
(that has $1$ at position $p$) and all of the convolved entity vectors (each having $1$ at position $p$);
this results in a vector of length $|\tau|$ where the $p$th bit is set to $1$ and all other bits set to $0$.
Finally, \dotsum() sums the outcome of the resulting products to give us a score of $1$.

To show that $\phi(x) = 0$ when $x \notin \tau$, we prove the contrapositive, namely that
if $\phi(x) = 1$, then $x$ must be a fact in $\tau$.
We proceed by contradiction. Assume that there exists a tuple $x \notin \tau$ such that $\phi(x) = 1$.
This means that at the time of computing the element-wise product in the $\dotsum{}$ function, 
there was a position $j$ at which all input vectors to $\dotsum{}$
had a value of $1$.
This can happen only when 
(1) applying the convolution filter $w_j$ to each of the entities in $x$ produces a vector having $1$ at position $j$, and
(2) the embedding of relation $r \in x$ has $1$ at position $j$.

The first case can happen only if all entities of $x$ appear in the $j$th fact $f_j \in \tau$;
the second case happens only if relation $r \in x$ appears in $f_j$.
But if all entities of $x$ as well as its relation appear in fact $f_j$,
then $x \in \tau$, contradicting our assumption. 
Therefore, if $\phi(x) = 1$, then $x$ must be a fact in $\tau$.

To complete the proof, we consider the case when $|\tau| = 0$.
In this case, since there are no facts, all entities and relations are represented by zero-vectors of length $\maxar$. 
Then, for any tuple $x$, $\phi(x) = 0$.
This completes the proof.
\end{proof}

\begin{theorem}[Expressivity of \oursimple{}]\label{theorem:expressivity-sup2}
Let $\tau$ be a set of true tuples defined over entities $\entset$ and relations $\relset$, and let
$\maxar = \max_{r \in \relset}(|r|)$ be the maximum arity of the relations in $\relset$.
There exists an \oursimple{} model with embedding vectors of size $\max(\maxar |\tau|, \maxar)$ 
that assigns $1$ to
the tuples in $\tau$ and $0$ to others.
\end{theorem}

\begin{proof}
To prove the theorem, we follow the same strategy as Theorem 1 by showing an assignment of embedding values for each entity and relation
in $\tau$ such that the scoring function of \oursimple{} is as follows:
\[
  \phi(x) =
  \begin{cases}
                                  1 & \text{if $x \in \tau$} \\
                                  0 & \text{otherwise}
  \end{cases}
\]

We begin the proof by first describing the embeddings of each of the entities and relations in \oursimple{};
we then proceed to show that with such an embedding, \oursimple{} can represent any world accurately.

Let us first assume that $|\tau| > 0$.
We let each entity $e \in \entset$ be represented with a vector of length $\maxar |\tau|$ in which the $p$th block of $|\tau|$-bits indicates entities that appeared in position $p$ in the facts. In an entity embedding, the $i$th bit in the $j$th block is set to 1 if the entity appears in position $j$ in the $i$th fact.
Each relation $r \in \relset$ is then represented as a vector of length $\maxar * \tau$ whose $p$th bit 
is equal to $1$ if fact $f_p$ is defined on relation $r$, and $0$ otherwise.

Given any tuple $x$, we want to show that $\phi(x) = 1$ if $x \in \tau$ and $0$ otherwise.

First assume that $x=f_i$ is the $i$th fact in $\tau$
that is defined on relation $r$ and entities $e_1, e_2, \dots$, where $e_p$ is the entity at position $p$.
Shifting each $e_p$ results in 
a vector where the $i$th bit is equal to $1$
(See Figure~\ref{fig:expressivity2}).
Then, as a first step, function \dotsum{} computes the element-wise multiplication between the embedding of relation $r$ 
(that has $1$ at position $i$) and all of the shifted entity vectors (each having $1$ at position $i$).
Finally, \dotsum() sums the outcome of the resulting products to give us a score of~$1$.

To show that $\phi(x) = 0$ when $x \notin \tau$, we prove the contrapositive, namely that
if $\phi(x) = 1$, then $x$ must be a fact in $\tau$.
We proceed by contradiction. Assume that there exists a tuple $x \notin \tau$ such that $\phi(x) = 1$.
This means that at the time of computing the element-wise product in the $\dotsum{}$ function, 
there was a position $i$ at which all input vectors to $\dotsum{}$
had a value of $1$.
This can happen only when 
(1) applying the shift operation to each of the entities in $x$ produces a vector having $1$ at position $i$, and
(2) the embedding of relation $r \in x$ has $1$ at position $i$.

The first case can happen only if all entities of $x$ appear in the $i$th fact $f_i \in \tau$;
the second case happens only if relation $r \in x$ appears in $f_i$.
But if all entities and relations of $x$ appear in fact $f_i$,
then $x \in \tau$, contradicting our assumption. 
Therefore, if $\phi(x) = 1$, then $x$ must be a fact in $\tau$.

To complete the proof, we consider the case when $|\tau| = 0$.
In this case, since there are no facts, all entities and relations are represented by zero-vectors of length $\maxar$. 
Then, for any tuple $x$, $\phi(x) = 0$.
This completes the proof.
\end{proof}

\end{document}